\documentclass[11pt]{article}

\pdfoutput=1

\usepackage[final]{acl}
\usepackage{authblk}
\usepackage{placeins} 
\usepackage{times}
\usepackage{latexsym}
\usepackage{float}
\usepackage{pgfplots}
\usepgfplotslibrary{polar}
\pgfplotsset{compat=1.18}
\pgfplotsset{every axis/.append style={
    tick label style={font=\scriptsize},
    label style={font=\scriptsize},
    title style={font=\scriptsize},
    legend style={font=\scriptsize}
}}
\usepackage{multirow}
\usepackage{amsmath}
\usepackage[capitalize]{cleveref}
\usepackage{amsfonts}
\usepackage[T1]{fontenc}

\usepackage[utf8]{inputenc}

\usepackage{microtype}

\usepackage{inconsolata}

\usepackage{graphicx}

\usepackage[export]{adjustbox}

\usepackage{xspace}
\usepackage{multirow}
\usepackage{cancel} 
\usepackage{booktabs} 
\usepackage[normalem]{ulem}

\crefname{equation}{Eq.}{Eqs.}
\crefname{figure}{Fig.}{Figures}
\crefname{tabular}{Tab.}{Tabs.}
\crefname{table}{Tab.}{Tables}
\Crefname{table}{Table}{Tables}
\crefname{section}{Sec.}{Secs.}
\crefname{appendix}{App.}{Apps.}

\newcommand{\mamba}[0]{Mamba\xspace}
\newcommand{\mambab}[0]{Mamba-2\xspace}
\newcommand{\inproj}[0]{InProj\xspace}
\newcommand{\outproj}[0]{OutProj\xspace}

\newcommand{\headdim}[0]{P}

\newcommand{\dmodel}[0]{D}

\newcommand{\phimamba}[0]{Phi-Mamba-1.5B\xspace}

\newcommand{\hybridmamba}[0]{HLM-3B\xspace}
\newcommand{\smol}[0]{Smol-Mamba-1.9B\xspace}

\newcommand{\R}{\mathbb{R}}

\newcommand{\numheads}[0]{H}
\newcommand{\statedim}[0]{N}

\newcommand{\wx}[0]{$W_X$}
\newcommand{\wz}[0]{$W_Z$}
\newcommand{\wb}[0]{$W_B$}
\newcommand{\wc}[0]{$W_C$}
\newcommand{\wa}[0]{$W_A$}
\newcommand{\wdee}[0]{$W_D$}
\newcommand{\wdelta}[0]{$W_{\Delta}$}

\definecolor{darkbrown}{HTML}{6A3717}
\definecolor{orange}{RGB}{255,128,0}

\title{On Pruning State-Space LLMs}

\author{\textbf{Tamer Ghattas} \qquad \textbf{Michael Hassid} \qquad \textbf{Roy Schwartz} \\
  The Hebrew University of Jerusalem \\
  \texttt{\{tamer.ghattas, michael.hassid, roy.schwartz1\}@mail.huji.ac.il}
  }

\begin{document}
\maketitle
\begin{abstract}

Recent work proposed state-space models~(SSMs) as an efficient alternative to transformer-based LLMs. Can these models be pruned to further reduce their computation costs?
We adapt several pruning methods to the SSM structure, and apply them to four SSM-based LLMs across multiple tasks. We find that such models are quite robust to some pruning methods (e.g., WANDA), while using other methods lead to fast performance degradation.\footnote{\url{www.github.com/schwartz-lab-NLP/SSM-Pruner}}

\end{abstract}

\begin{figure}[!t]  
    \centering
    \includegraphics[width=\linewidth]{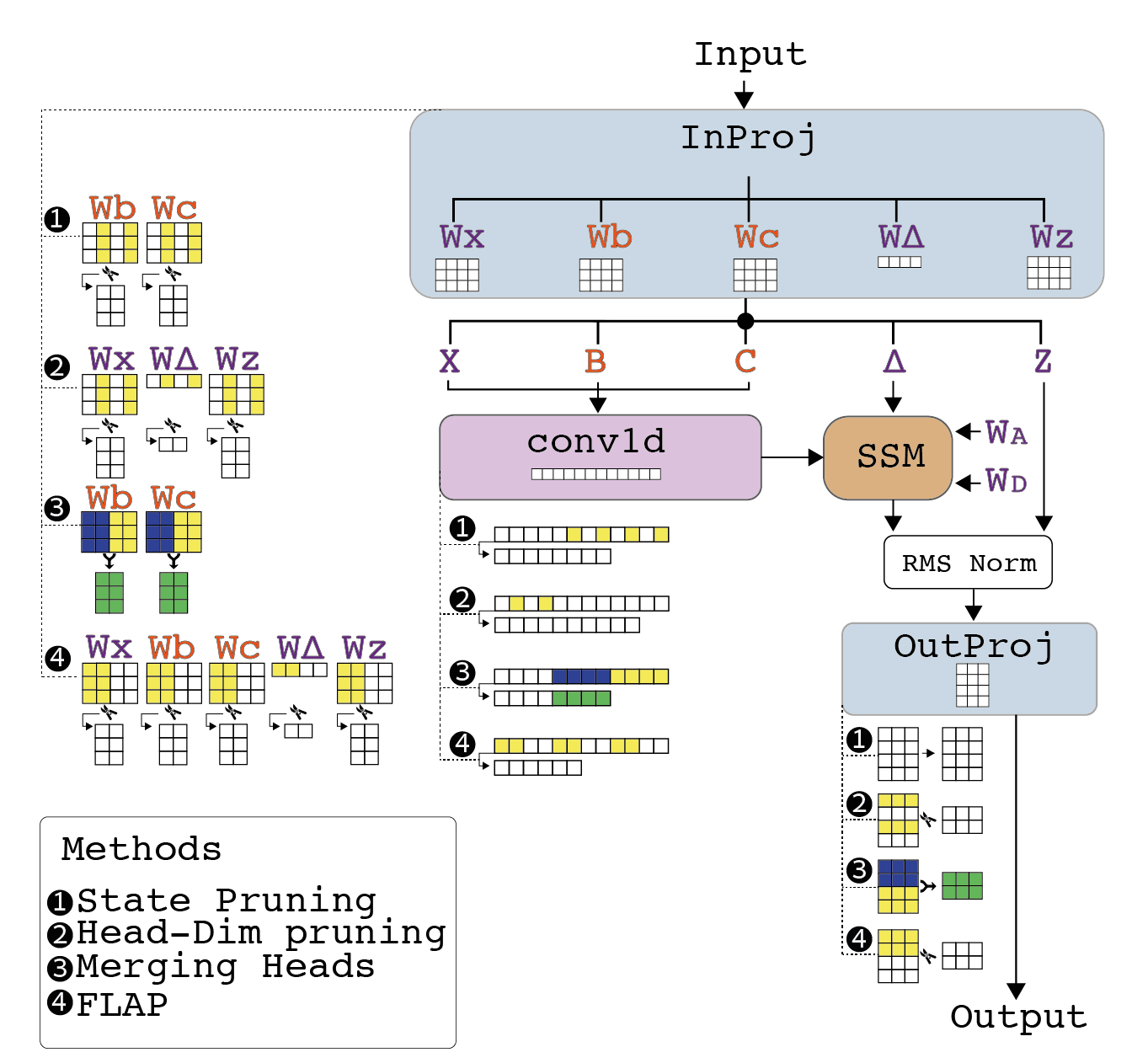}  
    \caption{Pruning SSM-based LLMs. \textbf{Right}: the \mamba SSM block: the input is linearly projected using five projection matrices~(\wz,\wx,\wb,\wc,\wdelta), to be used in later parts of the block.
    Every SSM head is represented using two vectors~(two rows for \inproj and two columns for \outproj). 
    \textbf{Left}: our different structure pruning methods. Each \textcolor{yellow!50!black}{yellow} cell represents a pruned element in the corresponding head.
    (1)~\textit{State pruning}: head extraction from \wb~and \wc~tensors then pruning the corresponding conv1d filters; 
    (2)~\textit{Head dimension pruning}: 
    head extraction from \wx, \wz, \wa, \wdee~and \wdelta, and pruning  the corresponding conv1d filters and \outproj rows;
    (3)~\textit{Head Merging}: mean-pooling every two BC-heads and all corresponding components; 
    (4)~\textit{SSM-FLAP}:
    adapting FLAP to SSMs, which prunes whole heads on all \inproj sub-components heads and their correspondingly conv1d and \outproj.}
    \label{fig:mamba2arch}
\end{figure}

\section{Introduction}
Selective-State Space models~(SSMs,~\citealp{gu2022s4}) have recently gained attention as an appealing alternative to transformers~\cite{mamba-1,mamba-2}. SSMs leverage both selective memory capabilities and RNN~\cite{RNN} properties, showing comparable results against transformer-based peers. However, SSM-based LLMs are still parameter-heavy, which raises the question of how well they can be compressed.

In this work, we focus on one of the key compression methods---\textit{pruning}~\cite{lecun1990pruning}. 
Modern LLM pruning methods have been developed and tested mostly for transformer components such as self-attention and feed-forward. Here we study how well SSM-based LLMs can be pruned.

We adapt several structured pruning methods to SSMs, e.g., pruning different SSM heads using different criteria, or merging existing heads~(\cref{fig:mamba2arch}). We apply these methods to four SSM-based LLMs, along with WANDA~\cite{wanda}, an unstructured pruning method that requires no adaptation. We compare all methods across six different tasks.

Our results show that all models are robust to unstructured pruning with WANDA, even when reducing up to 50\% of the SSM parameter count. 
We also observe that pruning SSM states leads to small degradation in almost all cases. In contrast, pruning the SSM heads leads to a sharp drop in performance in all cases. We also show that the output projection in SSM-based LLMs is much more sensitive to pruning than the input projection. Our results hint that SSM-based LLMs can indeed be made more efficient, but the choice of pruning method has a large effect on the pruning quality.

\section{Background}\label{sec:background}
In this section, we discuss recent SSM architectures and some of the best performing pruning methods commonly used with transformer-based models.
\subsection{SSM Architectures}
State Space Models~~\citep{gu2022s4} are a class of seq2seq models that represent inputs via hidden states evolving over time. Unlike attention-based architectures, SSMs leverage structured representations to achieve sub-quadratic complexity. 

Recent work has shown that SSM-based LLMs can be trained and reach competitive performance to transformers-based LLMs.
\mamba~\citep{mamba-1} uses a “selective” SSM variant that adaptively focuses on certain parts of the input at each time step.  
\mambab~\cite{mamba-2} builds upon structured space duality framework, which bridges the gap between recurrent-style processing and attention-like operations. This enables the use of hardware optimizations made for attention. By producing SSM parameters in parallel and simplifying its core layer, \mambab is 2–8× faster than its predecessor while maintaining performance on par with transformers across various benchmarks.

The SSM component in \mambab is composed of several sub-components~(\cref{fig:mamba2arch}). The input is first passed through a linear layer (\inproj), composed of the following tensors: \wx,\wz~$\in \R^{\dmodel, \numheads, \headdim}$; \wb, \wc~$\in \R^{\dmodel, \numheads,\statedim}$; \wdelta~$\in \R^{\dmodel,\numheads}$. This  results in five matrices: $X,B,C,\Delta,$ and $Z$.\footnote{Where $\dmodel$ is the model dimension, $\numheads$ is the number of heads, $\headdim$ is the head dimension, and $\statedim$ is the state dimension.}
Then, $X, B$, and $C$ are  passed through a 1-D convolution layer. Its output, along with \wdelta~and two learned parameter matrices \wa , \wdee $\in \R^{1,\numheads}$, are passed to the SSD algorithm~\cite{mamba-2}. Its output is then joint with \wz~and normalized using RMS Norm \cite{rms}, and projected back to $\dmodel$---the  model dimension.

\textbf{Multi-head patterns in SSM-based LLMs} 
Similarly to group-query attention in transformers~(GQA;~\citealp{GQAGoogle}), SSM-based LLMs allow grouping some of the SSM subcomponents. The choice of which components to merge is referred to as the \textit{multi-head pattern}~\cite{mamba-2}. In contrast to the commonly used GQA pattern in transformers, different SSM-based LLMs use different patterns. For example, \mambab~\cite{mamba-2} groups the \wb,\wc~matrices, while Hybrid-Llama3-Mamba2-3B (\hybridmamba;~\citealp{mambaInLlama}) groups \wx~and \wb.

\subsection{Pruning Methods}
Pruning is frequently used to compress LLMs. Below we describe common pruning methods. 

\paragraph{Unstructured pruning} induces sparsity in the model weights. Such methods reduce model size, though it is hard to translate this sparsity to runtime gains. A common implementation of this approach is \textit{magnitude pruning}, which prunes unimportant parameters based on their absolute values~\cite{frankle2019lth}. A recent highly effective variant of magnitude pruning is WANDA~\cite{wanda}, which also takes activations into account. Importantly, WANDA does not require fine-tuning, making it highly efficient for compressing LLMs.

\paragraph{Structured pruning} removes entire sections of model weights rather than individual elements~\cite{ma2023llmpruner,molchanov2019importanceestimationneuralnetwork,fan2019reducingtransformerdepthdemand}. 
One such method is FLAP~\cite{FLAP}, which prunes based on the fluctuation of each input channel. It determines whether the output feature map can be recovered after pruning a column from the weight matrix. FLAP avoids the need for retraining and requires only a single forward pass for both pruning and bias compensation.
Finally, group-query attention (GQA), which merges distinct attention KV heads using mean-pooling, can also be thought of as a structured pruning method.

\section{Pruning SSM-based LLMs}

We aim to study the effect of different pruning methods on SSM-based LLMs. Below we describe how we adapt different pruning methods, developed for transformers, to SSMs.
We note that within the \mambab SSM component~(\cref{fig:mamba2arch}), the layer dimension dictates the shapes for the matrices within the input projection (\inproj), which comprise approximately 67\% of all parameters in the SSM component. The output projection (\outproj) and the 1d convlution layer (conv1d) are roughly 32\% and < 1\%, respectively. 

\paragraph{WANDA} is an \textbf{unstructured pruning} method, which can be applied to SSM layers with minimal adjustments, as it originally operates on linear layers of any size without further restrictions. 

We also adapt four \textbf{structural pruning methods} to SSMs.

\paragraph{State pruning}
Each head in the \wb~and \wc~matrices of \mambab is composed of a $\dmodel \times \statedim$ matrix. Therefore,  
we can prune the least important tensors according to a desired ratio. In our experiments, we use the second order Taylor approximation-based importance estimator~\cite{molchanov2019importanceestimationneuralnetwork}, and then average this score per tensor in each head to estimate its importance. To do so, we perform model pass on 20 wikitext2 samples~\cite{merity2016pointersentinelmixturemodels} and accumulate gradients to calculate importance. To preserve the dimensionality correctness within the rest of the flow, we prune the \outproj matrix and the conv1d layer weights accordingly.    

\paragraph{Head Dimension Pruning}
Similarly, each head in the \wx~tensor is composed of $\headdim$ matrices, and thus can be pruned along with \wz~in the same way.

\paragraph{Merging Heads}
Inspired by the grouping of KV heads in attention~\cite{GQAGoogle,ConvertingMHAtoGQA}, we consider pruning SSMs by grouping the $BC$ or $XB$ heads, while maintaining the multi-head pattern, e.g., further merging the already grouped heads~(see~\cref{sec:background}). To do so, we use mean pooling on consecutive heads. An exception to the pattern preservation policy is \mambab, as it has only a single $BC$ head, so we merge its $X$ heads.

\paragraph{SSM-FLAP}
We adapt FLAP to prune SSM-based LLMs by applying its calculated pruning masks to the submatrices of \inproj and pruning the corresponding elements in the rest of the flow. To enable bias compensation, we add a bias term similar to how FLAP operates with attention. Since FLAP prunes a different number of heads per layer, in an MHA based model where the number of $BC$ heads and $X$ heads is the same, we prune both groups, while in a GVA based model, we exclude the $BC$ heads and limit the pruning to round the number of kept $X$ heads to the closest larger multiplier of $BC$ heads, keeping the number of $X$ heads dividable by the number of $BC$ heads.

\begin{table*}[!t]
\centering
\small
\renewcommand{\arraystretch}{1}

\begin{tabular}{|l|c|c|ccc|ccc|cc|}
\toprule
\multirow{2}{*}{\textbf{Model}} & \multirow{2}{*}{\textbf{Ratio}}  & \multicolumn{1}{c|}{\textbf{WANDA}} & \multicolumn{3}{c|}{\textbf{State}} & \multicolumn{3}{c|}{\textbf{Head}} & \multicolumn{2}{c|}{\textbf{SSM-FLAP}} \\
 & & w/o FT & w/o FT & w/ FT & Comp. & w/o FT & w/ FT & Comp. & w/o FT & Comp. \\
\midrule
\multirow{3}{*}{\shortstack[l]{Phi-Mamba-1.5B\\(SSM \& MLPs)}}
 & Dense 
   & 0.59 
   & 0.59  & N/A 
   & \phantom{0}0\% 
   & 0.59  & N/A 
   & \phantom{0}0\% 
   & 0.59 & \phantom{0}0\% \\
 & 25\%  
   & 0.59 
   & 0.58     & 0.59 
   & 10\% 
   & 0.40  & 0.54 
   & 25\% 
   & 0.57 & 26\% \\
 & 50\%  
   & 0.57 
   & 0.54  & 0.58 
   & 20\% 
   & 0.33  & 0.47 
   & 50\% 
   & 0.49 & 51\% \\
\midrule
\multirow{3}{*}{\shortstack[l]{\hybridmamba\\(SSM \& MLPs)}}
 & Dense  
   & 0.64 
   & 0.64  & N/A 
   & \phantom{0}0\% 
   & 0.64  & N/A 
   & \phantom{0}0\% 
   & 0.64 & \phantom{0}0\% \\
 & 25\%  
   & 0.64 
   & 0.30     & 0.31 
   & 20\% 
   & 0.30  & 0.32 
   & \phantom{0}5\% 
   & 0.29 & \phantom{0}5\% \\
 & 50\%  
   & 0.63 
   & 0.29  & 0.30 
   & 41\% 
   & 0.29  & 0.31 
   & 10\% 
   & 0.29 & 10\% \\
\midrule
\multirow{3}{*}{\shortstack[l]{\smol\\(SSM \& MLPs)}}
 & Dense  
   & 0.61 
   & 0.61  & N/A 
   & \phantom{0}0\% 
   & 0.61  & N/A 
   & \phantom{0}0\% 
   & 0.61 & \phantom{0}0\% \\
 & 25\%  
   & 0.60 
   & 0.59     & 0.60 
   & 13\% 
   & 0.29  & 0.30 
   & \phantom{0}5\% 
   & 0.51 & 26\% \\
 & 50\%  
   & 0.56 
   & 0.47  & 0.59 
   & 25\% 
   & 0.28  & 0.30 
   & 10\% 
   & 0.41 & 49\% \\
   \midrule
\multirow{3}{*}{\shortstack[l]{\mambab-2.7B\\(only SSM)}}
 & Dense  
   & 0.60 
   & 0.60  & N/A 
   & \phantom{0}0\% 
   & 0.60  & N/A 
   & \phantom{0}0\% 
   & 0.60 & \phantom{0}0\% \\
 & 25\%  
   & 0.53 
   & 0.53     & 0.54 
   & 0.5\% 
   & 0.29  & 0.30 
   & 24\% 
   & 0.30 & 25\% \\
 & 50\%  
   & 0.33 
   & 0.47  & 0.48 
   & \phantom{0}1\% 
   & 0.29  & 0.29 
   & 47\% 
   & 0.30 & 49\% \\
\bottomrule
\end{tabular}
\caption{Results for pruning SSM components in different ratios, with \textit{Dense} being an unpruned baseline. The State, Head, and SSM-FLAP methods report their SSM component compression (Comp.) values, along with the average benchmark accuracy before (w/o FT) and after (w/ FT) fine-tuning. Results for WANDA and FLAP are only w/o FT, as that they do not require fine-tuning to work well in practice. “N/A” denotes that no fine-tuning is performed.}
\label{tab:ratio_pruning}
\end{table*}

\begin{table}[!t]
\centering
 \small
\renewcommand{\arraystretch}{1}
\setlength{\tabcolsep}{2pt}
\begin{tabular}{|l|c|c|cc|}
\toprule
\textbf{Model} & \textbf{Heads} & \textbf{Comp.} & \textbf{w/o FT} & \textbf{w/ FT} \\
\midrule
\multirow{3}{*}{\shortstack[l]{\phimamba\\(SSM \& MLPs)}}
 & 32  & \phantom{0}0\%  & 0.59 & N/A \\
 & 16  & 20\% & 0.34 & 0.54 \\
 & 8   & 30\% & 0.31 & 0.48 \\
\midrule
\multirow{3}{*}{\shortstack[l]{\hybridmamba\\(SSM \& MLPs)}}
 & 8   & \phantom{0}0\%  & 0.64 & N/A \\
 & 4   & 10\% & 0.30 & 0.31 \\
 & 2   & 14\% & 0.29 & 0.30 \\
\midrule
\multirow{3}{*}{\shortstack[l]{\smol\\(SSM \& MLPs)}}
 & 32   & \phantom{0}0\%  & 0.61 & N/A \\
 & 16   & 25\% & 0.29 & 0.44 \\
 & 8   & 38\% & 0.29 & 0.41 \\
 \midrule
\multirow{3}{*}{\shortstack[l]{\mambab-2.7B\\(only SSM)}}
 & 80  & \phantom{0}0\%  & 0.60 & N/A \\
 & 40  & 49\% & 0.29 & 0.29 \\
 & 20  & 70\% & 0.28 & 0.28 \\
\bottomrule
\end{tabular}
\caption{Merging heads pruning results. Heads is the number of heads retained, Comp.~is the SSM component compression rate. The topline is the first row per model.}
\label{tab:heads_pruning}
\end{table}

\section{Experiments}

\paragraph{Models} There aren't many competitive SSM-based LLMs. We consider four main models in different sizes and base architectures:
\textsc{\mambab-2.7B}, which is configured with a multi-value attention head pattern; \textsc{\phimamba}~\cite{mohawk}, which is distilled from \textsc{Phi-1.5}~\cite{phi}. It preserves the same MLP, embeddings and LM head layers and converts the attention layer to be an SSM component. This model is configured with grouped-value attention; Hybrid-Llama3-Mamba2-3B~(\textsc{\hybridmamba};~\citealp{mambaInLlama}) is a hybrid model of interleaving attention and SSM layers distilled from \textsc{Llama-3.1-70B-Instruct}~\cite{grattafiori2024llama3herdmodels} but initialized using \textsc{Llama-3.2-3B}~\cite{grattafiori2024llama3herdmodels} weights. It converts only the attention layer but finetunes all model layers. Unlike the original architecture of \mambab, this model uses grouped query attention (GQA) since its initializing weights come from LLama, which is GQA based; 
Finally, we distill \textsc{\smol}, \footnote{We release the model at \url{https://huggingface.co/schwartz-lab/Smol2-Mamba-1.9B}.}, 
a Mamba model from \textsc{Smol2-1.7B}~\cite{smol}, using the MOHAWK method~\cite{mohawk}.
We note that \mambab is a pure SSM LLM. That is, it only contains SSM blocks. In contrast, the other three models contain  interleaving SSM and FFN layers.

\paragraph{Experimental setup} For WANDA, state, head, and SSM-FLAP pruning, we prune models by 25\% and 50\%. For merging heads, we merge 50\% and 75\% of the heads. In all cases we report the topline---the unpruned models. Pruned models are finetuned on wikitext2. We use LORA~\cite{hu2021loralowrankadaptationlarge} targeting both the SSM and MLP layers. We use the KD loss with the teacher set as the original model pre-pruning.\footnote{Preliminary experiments show that it outperforms standard CE loss, see~\cref{tab:CEvsKD} in \cref{sec:appendix}.} For each setup, we also report the ratio of pruned parameters of the full SSM component. See~\cref{app:exp_setup} for more details.
  
\paragraph{Benchmarks} We use the EleutherAI LM Harness\footnote{\url{www.github.com/EleutherAI/lm-evaluation-harness}} to experiment with lambada~\cite{paperno-etal-2016-lambada}, hellaswag~\cite{zellers2019hellaswag}, piqa~\cite{Bisk2020}, arc-easy~\cite{clark2018thinksolvedquestionanswering}, arc-challenge~\cite{clark2018thinksolvedquestionanswering}, and winogrande~\cite{sakaguchi2019winograndeadversarialwinogradschema}.

\paragraph{Results}
\Cref{tab:ratio_pruning} shows our pruning results for all methods except head merging~(shown in \cref{tab:heads_pruning}), averaged across tasks.\footnote{See~\cref{app:all_results} for full results on all tasks.}  
WANDA tends to preserve model quality across models, especially at moderate pruning (25\%) in 3/4 models, and does not collapse even at 50\%. The exception is \mambab-2.7B, which drops more quickly. This is somewhat expected, as in that case there are no FFN layers.\footnote{We also compare the effect of WANDA pruning on a similar transformer-based model~(\cref{app:pruning_advantage_checks}, \cref{tab:wanda_vs_tr}), observing a similar drop in performance for both.}

Our structured pruning approaches show larger variance across models. \textsc{\phimamba} maintains reasonable performance across state, head and SSM-FLAP pruning, even at 50\% ratios. \textsc{\smol} also performs well across most methods, except head pruning.
In contrast, \textsc{\hybridmamba} and \textsc{\mambab-2.7B} suffer severe degradations in almost all cases, even at moderate pruning ratios~(25\%) and post-finetuning.

\paragraph{Analysis}

We study whether \textsc{\mambab-2.7B}, the most sensitive to pruning,  exhibits different behavior to pruning different components. 
To do so, we apply WANDA exclusively on \inproj, \outproj, or both, and report perplexity on wikitext2. Our results~(\cref{fig:perplexity_plot}) show that pruning \outproj yields a drastic spike in perplexity, even at moderate pruning ratios (30–40\%), while \inproj can be pruned more aggressively without catastrophic effects. Importantly, this is despite \inproj having more than twice the number of parameters of \outproj.

To estimate the advantage of structural pruning, we measure the decoding throughput (tokens/second) of \smol before and after applying FLAP in \cref{app:pruning_advantage_checks}, \cref{tab:flap_throughput}. Pruning 50\% produces a speedup of 5--17\%, while pruning 25\% produces a speedup of 0--6\%.

Aiming to find the breaking point of WANDA pruning, we sweep the pruning ratio 0--90\% on \smol and report results on the same benchmarks in \cref{tab:wanda_ratio_sweep_smol}. The steepest drop occurs between 70\% and 80\%, where model performance drops substantially on several datasets, most notably LAMBADA.

\begin{table*}[t]
\centering
\footnotesize
\setlength{\tabcolsep}{6pt}
\renewcommand{\arraystretch}{1.05}
\begin{tabular}{lccccccc}
\toprule
\textbf{Ratio} & \textbf{ARC-C} & \textbf{ARC-E} & \textbf{HellaSwag} & \textbf{PIQA} & \textbf{Winogrande} & \textbf{LAMBADA} & \textbf{Average} \\
\midrule
0.00 & 0.430 & 0.750 & 0.510 & 0.770 & 0.630 & 0.570 & \textbf{0.610} \\
0.25 & 0.420 & 0.750 & 0.500 & 0.770 & 0.630 & 0.560 & \textbf{0.605} \\
0.50 & 0.370 & 0.720 & 0.470 & 0.760 & 0.590 & 0.420 & \textbf{0.555} \\
0.60 & 0.379 & 0.658 & 0.621 & 0.748 & 0.580 & 0.370 & \textbf{0.559} \\
0.70 & 0.328 & 0.587 & 0.555 & 0.717 & 0.526 & 0.195 & \textbf{0.485} \\
0.80 & 0.260 & 0.446 & 0.421 & 0.648 & 0.500 & 0.027 & \textbf{0.384} \\
0.90 & 0.239 & 0.320 & 0.280 & 0.547 & 0.491 & 0.000 & \textbf{0.313} \\
\bottomrule
\end{tabular}
\caption{Higher sparsity exploration for \smol\ with WANDA. Pruning ratio sweep between 0--0.9 in steps of 0.1. 
Performance drops most significantly between 0.7 and 0.8.
}
\label{tab:wanda_ratio_sweep_smol}
\end{table*}

\paragraph{Takeaways}
SSM-based LLMs seem robust to WANDA pruning.
Among structured pruning methods, state pruning seems most effective, leading to small to negligible performance drop in 3/4 models. 
In contrast, all models crash when applying head pruning, even at moderate rates. The other two methods (SSM-FLAP and head merging) work for some models but not others.

\section{Conclusions}
We adapted different pruning methods for SSM-based LLMs. Our results show that such LLMs can be pruned successfully with unstructured methods~(WANDA), or even structured 
ones~(state-pruning) with little to no performance degradation in some cases. Our results hint at the potential of making SSM-based LLMs even more efficient.

\begin{figure}
\centering
\begin{tikzpicture}
\small
    \begin{axis}[
        width=.45\textwidth, height=5cm,
        xlabel={Whole model sparsity (ratio)},
        ylabel={Perplexity},
        legend pos=south east,
        grid=both,
        xmin=0.0, xmax=0.7,
        ymin=0., ymax=2000,
        ymode=log,
        log basis y=10,
        xtick={0, 0.1,  0.2,  0.3,  0.4, 0.5},
        ytick={1, 10, 100, 1000, 2000},
    ]
        \addplot[
            color=blue,
            mark=o,
            thick
        ] coordinates {
            (0.3369, 10.5762)
            (0.2695, 9.4666)
            (0.2021, 9.0830)
            (0.1348, 8.9316)
        };
        \addlegendentry{\small \inproj}

        \node at (axis cs:0.3369,10.5762) [anchor=south west, blue] {50\%};
        \node at (axis cs:0.2695,9.4666) [anchor=south west, blue] {40\%};
        \node at (axis cs:0.2021,9.0830) [anchor=south west, blue] {30\%};
        \node at (axis cs:0.1348,8.9316) [anchor=south west, blue] {20\%};

        \addplot[
            color=red,
            mark=triangle,
            thick
        ] coordinates {
            (0.1631, 948.7922)
            (0.1305, 141.4194)
            (0.0979, 33.3519)
            (0.0652, 18.7741)
        };
        \addlegendentry{\small \outproj}

        \node at (axis cs:0.1631,948.7922) [anchor=south east, red] {50\%};
        \node at (axis cs:0.1305,141.4194) [anchor=south east, red] {40\%};
        \node at (axis cs:0.0979,33.3519) [anchor=south east, red] {30\%};
        \node at (axis cs:0.0652,18.7741) [anchor=south east, red] {20\%};

        \addplot[
            color=green,
            mark=square,
            thick
        ] coordinates {
            (0.5000, 1639.3659)
            (0.4000, 157.2813)
            (0.3000, 34.7289)
            (0.2000, 18.9848)
        };
        \addlegendentry{\small Both}

        \node at (axis cs:0.5000,1639.3659) [anchor=south west, green] {50\%};
        \node at (axis cs:0.4000,157.2813) [anchor=south west, green] {40\%};
        \node at (axis cs:0.3000,34.7289) [anchor=south west, green] {30\%};
        \node at (axis cs:0.2000,18.9848) [anchor=south west, green] {20\%};

    \end{axis}
\end{tikzpicture}

\caption{The effect of WANDA pruning ratios on different \textsc{\mambab-2.7B} components. 
OutProj layer is substantially more sensitive to pruning than InProj.}
\label{fig:perplexity_plot}
\end{figure}


\section*{Limitations}
Drawing direct conclusions from our experiments is not straightforward. The four LLMs we experiment with differ, among other, in size, head pattern, training data, and structure. These choices are due to the scarcity of SSM-based LLMs, and the computational costs of training or distilling various models to control for specific variables.

In addition, the structure pruning methods we consider also differ in their effect on the models, as indicated by the different compression ratios in \cref{tab:ratio_pruning,tab:heads_pruning}. The challenges here stem from the different constraints imposed by the different methods. E.g., in state pruning we prune $BC$-heads, which cap the compression by the number of parameters \wb~and \wc~occupy in the \mambab component, especially when the model configuration is GVA which translates to fewer $BC$-heads. Another example is when pruning Head-Dim we are obliged to prune large portions of \outproj to keep dimensionality correctness in the flow, leading to pruning potentially important parameters from it.

Despite these issues, we believe the signals we observe are valuable. For instance, the robustness of SSM-based LLMs to state pruning is demonstrated by the modest performance drop for 50\% pruning ratios~(e.g., a 1\% drop both with a 20\% compression ratio for in \textsc{\phimamba}, and 25\% compression ratio for \textsc{\smol}), compared to a huge drop with head pruning for a 25\% pruning ratio, sometimes with the same model~(e.g., a 30\% drop for both \textsc{\hybridmamba} and \textsc{\smol} with 5\% compression ratio).

We also focused our experiments exclusively on pruning the \mambab components and excluded feed-forward networks, assuming that prior work on transformer pruning had already addressed those extensively. Future work could address pruning all model components to study the interaction between the different components, as well as potential to get more significant computational savings. 

\section*{Acknowledgments}
We thank Jana Omary Ghattas for her help in creating \cref{fig:mamba2arch}. We also thank Yuval Reif for his support. 
This work was supported in part by the Israel Science Foundation (grant no. 2045/21).

\bibliography{custom, anthology}

\begin{thebibliography}{27}
\providecommand{\natexlab}[1]{#1}

\bibitem[{Ainslie et~al.(2023)Ainslie, Lee-Thorp, de~Jong, Zemlyanskiy, Lebron, and Sanghai}]{GQAGoogle}
Joshua Ainslie, James Lee-Thorp, Michiel de~Jong, Yury Zemlyanskiy, Federico Lebron, and Sumit Sanghai. 2023.
\newblock \href {https://doi.org/10.18653/v1/2023.emnlp-main.298} {{GQA}: Training generalized multi-query transformer models from multi-head checkpoints}.
\newblock In \emph{Proceedings of the 2023 Conference on Empirical Methods in Natural Language Processing}, pages 4895--4901, Singapore. Association for Computational Linguistics.

\bibitem[{Allal et~al.(2025)Allal, Lozhkov, Bakouch, Blázquez, Penedo, Tunstall, Marafioti, Kydlíček, Lajarín, Srivastav, Lochner, Fahlgren, Nguyen, Fourrier, Burtenshaw, Larcher, Zhao, Zakka, Morlon, Raffel, von Werra, and Wolf}]{smol}
Loubna~Ben Allal, Anton Lozhkov, Elie Bakouch, Gabriel~Martín Blázquez, Guilherme Penedo, Lewis Tunstall, Andrés Marafioti, Hynek Kydlíček, Agustín~Piqueres Lajarín, Vaibhav Srivastav, Joshua Lochner, Caleb Fahlgren, Xuan-Son Nguyen, Clémentine Fourrier, Ben Burtenshaw, Hugo Larcher, Haojun Zhao, Cyril Zakka, Mathieu Morlon, Colin Raffel, Leandro von Werra, and Thomas Wolf. 2025.
\newblock \href {https://arxiv.org/abs/2502.02737} {Smollm2: When smol goes big -- data-centric training of a small language model}.
\newblock \emph{Preprint}, arXiv:2502.02737.

\bibitem[{An et~al.(2023)An, Zhao, Yu, Tang, and Wang}]{FLAP}
Yongqi An, Xu~Zhao, Tao Yu, Ming Tang, and Jinqiao Wang. 2023.
\newblock \href {https://arxiv.org/abs/2312.11983} {Fluctuation-based adaptive structured pruning for large language models}.
\newblock \emph{Preprint}, arXiv:2312.11983.

\bibitem[{Bick et~al.(2025)Bick, Katsch, Sohoni, Desai, and Gu}]{bick2025llambascalingdistilledrecurrent}
Aviv Bick, Tobias Katsch, Nimit Sohoni, Arjun Desai, and Albert Gu. 2025.
\newblock \href {https://arxiv.org/abs/2502.14458} {Llamba: Scaling distilled recurrent models for efficient language processing}.
\newblock \emph{Preprint}, arXiv:2502.14458.

\bibitem[{Bick et~al.(2024)Bick, Li, Xing, Kolter, and Gu}]{mohawk}
Aviv Bick, Kevin~Y. Li, Eric~P. Xing, J.~Zico Kolter, and Albert Gu. 2024.
\newblock \href {https://arxiv.org/abs/2408.10189} {Transformers to ssms: Distilling quadratic knowledge to subquadratic models}.
\newblock In \emph{Proc. of NeurIPS}.

\bibitem[{Bisk et~al.(2020)Bisk, Zellers, Bras, Gao, and Choi}]{Bisk2020}
Yonatan Bisk, Rowan Zellers, Ronan~Le Bras, Jianfeng Gao, and Yejin Choi. 2020.
\newblock Piqa: Reasoning about physical commonsense in natural language.
\newblock In \emph{Thirty-Fourth AAAI Conference on Artificial Intelligence}.

\bibitem[{Clark et~al.(2018)Clark, Cowhey, Etzioni, Khot, Sabharwal, Schoenick, and Tafjord}]{clark2018thinksolvedquestionanswering}
Peter Clark, Isaac Cowhey, Oren Etzioni, Tushar Khot, Ashish Sabharwal, Carissa Schoenick, and Oyvind Tafjord. 2018.
\newblock \href {https://arxiv.org/abs/1803.05457} {Think you have solved question answering? try arc, the ai2 reasoning challenge}.
\newblock \emph{Preprint}, arXiv:1803.05457.

\bibitem[{Dao and Gu(2024)}]{mamba-2}
Tri Dao and Albert Gu. 2024.
\newblock \href {https://arxiv.org/abs/2405.21060} {Transformers are ssms: Generalized models and efficient algorithms through structured state space duality}.
\newblock \emph{Preprint}, arXiv:2405.21060.

\bibitem[{Elman(1990)}]{RNN}
Jeffrey~L. Elman. 1990.
\newblock \href {https://doi.org/10.1016/0364-0213(90)90002-E} {Finding structure in time}.
\newblock \emph{Cognitive Science}, 14(2):179--211.

\bibitem[{Fan et~al.(2019)Fan, Grave, and Joulin}]{fan2019reducingtransformerdepthdemand}
Angela Fan, Edouard Grave, and Armand Joulin. 2019.
\newblock \href {https://arxiv.org/abs/1909.11556} {Reducing transformer depth on demand with structured dropout}.
\newblock \emph{Preprint}, arXiv:1909.11556.

\bibitem[{Frankle and Carbin(2019)}]{frankle2019lth}
Jonathan Frankle and Michael Carbin. 2019.
\newblock \href {https://arxiv.org/abs/1803.03635} {The lottery ticket hypothesis: Finding sparse, trainable neural networks}.
\newblock In \emph{Proc. of ICLR}.

\bibitem[{Grattafiori et~al.(2024)Grattafiori, Dubey, Jauhri, Pandey, Kadian, Al-Dahle, Letman, Mathur, Schelten, Vaughan, Yang, Fan, Goyal, Hartshorn, Yang, Mitra, Sravankumar, Korenev, Hinsvark, and et~al.}]{grattafiori2024llama3herdmodels}
Aaron Grattafiori, Abhimanyu Dubey, Abhinav Jauhri, Abhinav Pandey, Abhishek Kadian, Ahmad Al-Dahle, Aiesha Letman, Akhil Mathur, Alan Schelten, Alex Vaughan, Amy Yang, Angela Fan, Anirudh Goyal, Anthony Hartshorn, Aobo Yang, Archi Mitra, Archie Sravankumar, Artem Korenev, Arthur Hinsvark, and Arun~Rao et~al. 2024.
\newblock \href {https://arxiv.org/abs/2407.21783} {The llama 3 herd of models}.
\newblock \emph{Preprint}, arXiv:2407.21783.

\bibitem[{Gu and Dao(2024)}]{mamba-1}
Albert Gu and Tri Dao. 2024.
\newblock \href {https://arxiv.org/abs/2312.00752} {Mamba: Linear-time sequence modeling with selective state spaces}.
\newblock {arXiv}:2312.00752.

\bibitem[{Gu et~al.(2022)Gu, Goel, and Ré}]{gu2022s4}
Albert Gu, Karan Goel, and Christopher Ré. 2022.
\newblock \href {https://arxiv.org/abs/2111.00396} {Efficiently modeling long sequences with structured state spaces}.
\newblock In \emph{Proc. of ICLR}.

\bibitem[{Hu et~al.(2021)Hu, Shen, Wallis, Allen-Zhu, Li, Wang, Wang, and Chen}]{hu2021loralowrankadaptationlarge}
Edward~J. Hu, Yelong Shen, Phillip Wallis, Zeyuan Allen-Zhu, Yuanzhi Li, Shean Wang, Lu~Wang, and Weizhu Chen. 2021.
\newblock \href {https://arxiv.org/abs/2106.09685} {Lora: Low-rank adaptation of large language models}.
\newblock \emph{Preprint}, arXiv:2106.09685.

\bibitem[{Jin et~al.(2024)Jin, Song, Zhou, and Qin}]{ConvertingMHAtoGQA}
Qingyun Jin, Xiaohui Song, Feng Zhou, and Zengchang Qin. 2024.
\newblock \href {https://arxiv.org/abs/2412.20677} {Align attention heads before merging them: An effective way for converting {MHA} to {GQA}}.
\newblock {arXiv}:2412.20677.

\bibitem[{LeCun et~al.(1989)LeCun, Denker, and Solla}]{lecun1990pruning}
Yann LeCun, John Denker, and Sara Solla. 1989.
\newblock \href {https://proceedings.neurips.cc/paper/1989/file/6c9882bbac1c7093bd25041881277658-Paper.pdf} {Optimal brain damage}.
\newblock In \emph{Advances in Neural Information Processing Systems}, volume~2. Morgan-Kaufmann.

\bibitem[{Li et~al.(2023)Li, Bubeck, Eldan, Giorno, Gunasekar, and Lee}]{phi}
Yuanzhi Li, Sébastien Bubeck, Ronen Eldan, Allie~Del Giorno, Suriya Gunasekar, and Yin~Tat Lee. 2023.
\newblock \href {https://arxiv.org/abs/2309.05463} {Textbooks are all you need ii: phi-1.5 technical report}.
\newblock \emph{Preprint}, arXiv:2309.05463.

\bibitem[{Ma et~al.(2023)Ma, Fang, and Wang}]{ma2023llmpruner}
Xinyin Ma, Gongfan Fang, and Xinchao Wang. 2023.
\newblock Llm-pruner: On the structural pruning of large language models.
\newblock In \emph{Advances in Neural Information Processing Systems}.

\bibitem[{Merity et~al.(2016)Merity, Xiong, Bradbury, and Socher}]{merity2016pointersentinelmixturemodels}
Stephen Merity, Caiming Xiong, James Bradbury, and Richard Socher. 2016.
\newblock \href {https://arxiv.org/abs/1609.07843} {Pointer sentinel mixture models}.
\newblock \emph{Preprint}, arXiv:1609.07843.

\bibitem[{Molchanov et~al.(2019)Molchanov, Mallya, Tyree, Frosio, and Kautz}]{molchanov2019importanceestimationneuralnetwork}
Pavlo Molchanov, Arun Mallya, Stephen Tyree, Iuri Frosio, and Jan Kautz. 2019.
\newblock \href {https://arxiv.org/abs/1906.10771} {Importance estimation for neural network pruning}.
\newblock \emph{Preprint}, arXiv:1906.10771.

\bibitem[{Paperno et~al.(2016)Paperno, Kruszewski, Lazaridou, Pham, Bernardi, Pezzelle, Baroni, Boleda, and Fern{\'a}ndez}]{paperno-etal-2016-lambada}
Denis Paperno, Germ{\'a}n Kruszewski, Angeliki Lazaridou, Ngoc~Quan Pham, Raffaella Bernardi, Sandro Pezzelle, Marco Baroni, Gemma Boleda, and Raquel Fern{\'a}ndez. 2016.
\newblock \href {https://doi.org/10.18653/v1/P16-1144} {The {LAMBADA} dataset: Word prediction requiring a broad discourse context}.
\newblock In \emph{Proceedings of the 54th Annual Meeting of the Association for Computational Linguistics (Volume 1: Long Papers)}, pages 1525--1534, Berlin, Germany. Association for Computational Linguistics.

\bibitem[{Sakaguchi et~al.(2019)Sakaguchi, Bras, Bhagavatula, and Choi}]{sakaguchi2019winograndeadversarialwinogradschema}
Keisuke Sakaguchi, Ronan~Le Bras, Chandra Bhagavatula, and Yejin Choi. 2019.
\newblock \href {https://arxiv.org/abs/1907.10641} {Winogrande: An adversarial winograd schema challenge at scale}.
\newblock \emph{Preprint}, arXiv:1907.10641.

\bibitem[{Sun et~al.(2024)Sun, Liu, Bair, and Kolter}]{wanda}
Mingjie Sun, Zhuang Liu, Anna Bair, and J.~Zico Kolter. 2024.
\newblock \href {https://arxiv.org/abs/2306.11695} {A simple and effective pruning approach for large language models}.
\newblock \emph{Preprint}, arXiv:2306.11695.

\bibitem[{Wang et~al.(2024)Wang, Paliotta, May, Rush, and Dao}]{mambaInLlama}
Junxiong Wang, Daniele Paliotta, Avner May, Alexander~M. Rush, and Tri Dao. 2024.
\newblock \href {https://arxiv.org/abs/2408.15237} {The mamba in the llama: Distilling and accelerating hybrid models}.
\newblock In \emph{Proc. of NeurIPS}.

\bibitem[{Zellers et~al.(2019)Zellers, Holtzman, Bisk, Farhadi, and Choi}]{zellers2019hellaswag}
Rowan Zellers, Ari Holtzman, Yonatan Bisk, Ali Farhadi, and Yejin Choi. 2019.
\newblock Hellaswag: Can a machine really finish your sentence?
\newblock In \emph{Proceedings of the 57th Annual Meeting of the Association for Computational Linguistics}.

\bibitem[{Zhang and Sennrich(2019)}]{rms}
Biao Zhang and Rico Sennrich. 2019.
\newblock \href {https://arxiv.org/abs/1910.07467} {Root mean square layer normalization}.
\newblock \emph{Preprint}, arXiv:1910.07467.

\end{thebibliography}

\newpage

\appendix
\crefalias{section}{appendix}
\section{Selecting the Loss Function}
Recovering performance in purned models is usually done by finetuning the model on a calibration dataset, however, this comes with the risk of over-fitting the dataset or regaining performance back in the area the dataset focuses on. Therefore, we checked the effect of using KD loss instead of Cross-Entropy~(CE) loss in the data set. As can be seen in~\cref{tab:CEvsKD}, it is clear that using KD loss helps the model regain overall performance along most benchmarks, while using CE, the model regains some performance in some tasks~(e.g., lambada) but doesn't improve much on other benchmarks.
\label{sec:appendix}
\begin{table*}
\centering
\small
\begin{tabular}{lccccccc}
\hline
{} &  arc\_challenge &  arc\_easy &  hellaswag &  lambada\_openai &   piqa &  winogrande &  Average \\
\hline
Baseline &         0.418 &    0.739 &     0.461 &          0.500 & 0.755 &      0.716 &   0.598 \\
w/o FT. &         0.378 &    0.713 &     0.433 &          0.349 & 0.747 &      0.651 &   0.545 \\
CE loss &         0.385 &    0.721 &     0.413 &          0.405 & 0.748 &      0.682 &   0.559 \\
KD loss &         0.410 &    0.726 &     0.454 &          0.452 & 0.753 &      0.693 &   0.581 \\
\hline
\end{tabular}
\caption{\label{tab:CEvsKD}
    Comparing loss choice. \phimamba with pruned SSM states according to Taylor importance estimation,  by 50\% 
 and fine-tuning using LoRA with different losses. The baseline model is the (full) fine-tuned model. KD loss consistently outperforms CE loss.
  }
\end{table*}

\section{Experimental Details}\label{app:exp_setup}
We finetune all models for 10K steps with batch size of 6 and learning rate of $5e-5$, using bf16. We use LoRA with rank=8 and $\alpha$=16.
We run our experiments on a single NVIDIA A100 80GB.

\section{Full Results}
\label{app:all_results}
\Cref{tab:headdim_full,tab:wanda_full,tab:merge_full,tab:baselines_full,tab:flap_results,tab:state_results} show our results on all benchmarks for the different models.

\section{MAMBA Pruning Advantage Experiments}
\label{app:pruning_advantage_checks}
\paragraph{Task accuracy under WANDA sparsity (\Cref{tab:wanda_vs_tr}).}
\Cref{tab:wanda_vs_tr} contrasts the accuracy of the transformer‐based Smol17 (TF) and its Mamba distilled version \smol when pruned with WANDA at 0 \%, 25 \%, and 50 \% sparsity.
At 25 \% sparsity, both architectures preserve their dense performance: the average score declines by only 0.3 pp for Smol17 (TF) and 0.4 pp for \smol, a relative drop below 1 \%.
Pruning 50 \% of the parameters yields a moderate degradation of 6.5 \% relative to the transformer and 8 \% for the Mamba variant.
These results indicate that WANDA can nullify up to 25 \% of weights from both model families with negligible impact, and as much as 50 \% while retaining useful accuracy, with a slightly larger advantage for the transformer based model.

\paragraph{End-to-end throughput with FLAP pruning (\Cref{tab:flap_throughput}).}
\Cref{tab:flap_throughput} reports batch inference throughput on a single NVIDIA A100-40 GB (host: 128 GB RAM, 128 CPU cores) when only Mamba blocks are pruned using FLAP.
With 50 \% sparsity (FLAP p0.5), throughput improves by 5–17 \% relative to the dense \smol baseline, the largest gains occurring for longer sequences and larger batch sizes (e.g., +17 \% at batch 16, seq 128).
At 25 \% sparsity (FLAP p0.25), the speed increases shrink to 0–6 \%, and the lightly pruned model occasionally falls slightly below baseline for the longest sequences, suggesting that mild sparsity does not offset the overhead of sparse kernels when memory bandwidth dominates.
To conclude, these findings demonstrate that aggressive but still accuracy-preserving FLAP pruning of Mamba components offers measurable end-to-end acceleration, particularly when high throughput is desired in a practical deployment.

\paragraph{LLAMBA 1B checks (\Cref{tab:flap_results_llamba} \& \Cref{tab:wanda_results_llamba}).}
In a concurrent work ~\cite{bick2025llambascalingdistilledrecurrent} to ours there was a release of additional family of Mamba distilled models from transformer base models, we ran pruning with WANDA and FLAP and ran the same benchmarks and put the results in \Cref{tab:wanda_results_llamba} and \Cref{tab:flap_results_llamba}.

\begin{table*}[ht]
\centering
\small
\begin{tabular}{lccccccccc}
\toprule
Model & Ratio & FT & arc\_challenge & arc\_easy & hellaswag & lambada & piqa & winogrande  \\
\midrule
\hybridmamba & 0.25 & w/o & 0.18 & 0.37 & 0.27 & 0.06 & 0.54 & 0.51  \\
\hybridmamba & 0.25 & w/  & 0.20 & 0.34 & 0.27 & 0.06 & 0.56 & 0.51  \\
\hybridmamba & 0.50 & w/o & 0.22 & 0.26 & 0.26 & 0.00 & 0.55 & 0.50  \\
\hybridmamba & 0.50 & w/  & 0.17 & 0.35 & 0.27 & 0.06 & 0.56 & 0.51  \\
\phimamba   & 0.50 & w/o & 0.18 & 0.40 & 0.27 & 0.00 & 0.60 & 0.52   \\
\phimamba   & 0.50 & w/  & 0.29 & 0.65 & 0.38 & 0.22 & 0.71 & 0.54   \\
\phimamba   & 0.25 & w/o & 0.25 & 0.60 & 0.30 & 0.06 & 0.69 & 0.50   \\
\phimamba   & 0.25 & w/  & 0.36 & 0.70 & 0.42 & 0.35 & 0.74 & 0.63   \\
\mambab-2.7B & 0.25 & w/o & 0.21 & 0.27 & 0.26 & 0.00 & 0.53 & 0.50  \\
\mambab-2.7B & 0.25 & w/  & 0.22 & 0.25 & 0.26 & 0.00 & 0.52 & 0.45  \\
\mambab-2.7B & 0.50 & w/o & 0.22 & 0.26 & 0.25 & 0.00 & 0.52 & 0.50  \\
\mambab-2.7B & 0.50 & w/  & 0.23 & 0.26 & 0.26 & 0.00 & 0.53 & 0.50  \\
\smol       & 0.25 & w/o & 0.22 & 0.25 & 0.26 & 0.00 & 0.52 & 0.49   \\
\smol       & 0.25 & w/  & 0.18 & 0.33 & 0.26 & 0.00 & 0.55 & 0.52   \\
\smol       & 0.50 & w/o & 0.20 & 0.26 & 0.26 & 0.00 & 0.50 & 0.49   \\
\smol       & 0.50 & w/  & 0.19 & 0.32 & 0.26 & 0.00 & 0.54 & 0.51   \\
\bottomrule
\end{tabular}
\caption{Head dimension pruning benchmarks results. See \Cref{fig:headdim_radar} for radar plot visualization.}
\label{tab:headdim_full}
\end{table*}

\begin{table*}[ht]
\centering
\small
\begin{tabular}{l c c c c c c c}
\toprule
Model         & Ratio & arc\_challenge & arc\_easy & hellaswag & lambada & piqa & winogrande \\
\midrule
\phimamba     & 0.25  & 0.41           & 0.74      & 0.46      & 0.50    & 0.76 & 0.72 \\
\phimamba     & 0.50  & 0.39           & 0.72      & 0.45      & 0.40    & 0.75 & 0.69 \\
\mambab-2.7B  & 0.25  & 0.31           & 0.63      & 0.42      & 0.49    & 0.72 & 0.58 \\
\mambab-2.7B  & 0.50  & 0.20           & 0.35      & 0.28      & 0.02    & 0.57 & 0.54 \\
\hybridmamba  & 0.25  & 0.51           & 0.80      & 0.55      & 0.55    & 0.77 & 0.68 \\
\hybridmamba  & 0.50  & 0.51           & 0.80      & 0.54      & 0.54    & 0.76 & 0.66 \\
\smol         & 0.25  & 0.42           & 0.75      & 0.50      & 0.56    & 0.77 & 0.63 \\
\smol         & 0.50  & 0.37           & 0.72      & 0.47      & 0.42    & 0.76 & 0.59 \\
\bottomrule
\end{tabular}
\caption{WANDA pruning benchmarks results. See \Cref{fig:wanda_radar} for radar plot visualization.}
\label{tab:wanda_full}
\end{table*}

\begin{table*}[ht]
\centering
\small
\begin{tabular}{l c c c c c c c c}
\toprule
model    & Heads & FT & arc\_challenge & arc\_easy & hellaswag & lambada & piqa & winogrande \\
\midrule
\phimamba         & 16  & w/o & 0.39 & 0.72 & 0.47 & 0.42 & 0.75 & 0.59 \\
\phimamba         & 16  & w/  & 0.45 & 0.73 & 0.49 & 0.43 & 0.76 & 0.62 \\
\phimamba         & 8   & w/o & 0.38 & 0.71 & 0.46 & 0.41 & 0.74 & 0.57 \\
\phimamba         & 8   & w/  & 0.41 & 0.72 & 0.47 & 0.42 & 0.75 & 0.58 \\
\mambab-2.7B      & 40  & w/o & 0.21 & 0.27 & 0.26 & 0.00 & 0.52 & 0.50 \\
\mambab-2.7B      & 40  & w/  & 0.21 & 0.27 & 0.25 & 0.00 & 0.52 & 0.50 \\
\mambab-2.7B      & 20  & w/o & 0.21 & 0.26 & 0.26 & 0.00 & 0.51 & 0.48 \\
\mambab-2.7B      & 20  & w/  & 0.21 & 0.26 & 0.26 & 0.00 & 0.52 & 0.48 \\
\smol             & 16  & w/o & 0.20 & 0.27 & 0.26 & 0.00 & 0.53 & 0.49 \\
\smol             & 16  & w/  & 0.27 & 0.59 & 0.41 & 0.14 & 0.70 & 0.53 \\
\smol             & 8   & w/o & 0.22 & 0.26 & 0.26 & 0.00 & 0.52 & 0.48 \\
\smol             & 8   & w/  & 0.32 & 0.66 & 0.40 & 0.25 & 0.72 & 0.54 \\
\hybridmamba      & 4   & w/o & 0.20 & 0.25 & 0.24 & 0.00 & 0.50 & 0.47 \\
\hybridmamba      & 4   & w/  & 0.22 & 0.27 & 0.26 & 0.00 & 0.53 & 0.50 \\
\hybridmamba      & 2   & w/o & 0.21 & 0.24 & 0.25 & 0.00 & 0.51 & 0.46 \\
\hybridmamba      & 2   & w/  & 0.23 & 0.28 & 0.27 & 0.00 & 0.54 & 0.49 \\
\bottomrule
\end{tabular}
\caption{Head merging benchmarks results. See \Cref{fig:merge_radar} for radar plot visualization.}
\label{tab:merge_full}
\end{table*}
       
\begin{table*}[ht]
\centering
\small
\begin{tabular}{lccccccc}
\toprule
model   & arc\_challenge & arc\_easy & hellaswag & lambada & piqa & winogrande \\
\midrule
\smol              & 0.43 & 0.75 & 0.51 & 0.57 & 0.77 & 0.63 \\
\mambab-2.7B       & 0.33 & 0.70 & 0.50 & 0.70 & 0.76 & 0.64 \\
\phimamba          & 0.41 & 0.74 & 0.46 & 0.50 & 0.76 & 0.72 \\
\hybridmamba       & 0.51 & 0.80 & 0.55 & 0.55 & 0.77 & 0.67 \\
\bottomrule
\end{tabular}
\caption{Models baseline results for the benchmarks.}
\label{tab:baselines_full}
\end{table*}

\clearpage
\begin{table*}[ht]
\centering
\small
\begin{tabular}{l c c c c c c c}
\toprule
\textbf{Model}    & \textbf{Ratio} & \textbf{arc\_challenge} & \textbf{arc\_easy} & \textbf{hellaswag} & \textbf{lambada} & \textbf{piqa} & \textbf{winogrande} \\
\midrule
\phimamba         & 0.25 & 0.39 & 0.72 & 0.44 & 0.44 & 0.75 & 0.66 \\
\phimamba         & 0.50 & 0.34 & 0.69 & 0.40 & 0.17 & 0.74 & 0.58 \\
\hybridmamba      & 0.25 & 0.22 & 0.27 & 0.26 & 0.00 & 0.52 & 0.49 \\
\hybridmamba      & 0.50 & 0.24 & 0.27 & 0.26 & 0.00 & 0.51 & 0.45 \\
\mambab-2.7B      & 0.25 & 0.23 & 0.28 & 0.25 & 0.00 & 0.51 & 0.48 \\
\mambab-2.7B      & 0.50 & 0.25 & 0.28 & 0.25 & 0.02 & 0.50 & 0.44 \\
\smol             & 0.25 & 0.34 & 0.69 & 0.44 & 0.27 & 0.74 & 0.55 \\
\smol             & 0.50 & 0.26 & 0.57 & 0.37 & 0.06 & 0.69 & 0.52 \\
\bottomrule
\end{tabular}
\caption{FLAP benchmark results. See \Cref{fig:flap_radar} for radar plot visualization.}
\label{tab:flap_results}
\end{table*}

\begin{table*}[ht]
\centering
\small
\begin{tabular}{l c c c c c c c c}
\toprule
\textbf{Model}    & \textbf{Ratio} & \textbf{FT} & \textbf{arc\_challenge} & \textbf{arc\_easy} & \textbf{hellaswag} & \textbf{lambada} & \textbf{piqa} & \textbf{winogrande} \\
\midrule
\hybridmamba  & 0.25 & w/o & 0.25 & 0.25 & 0.24 & 0.02 & 0.49 & 0.49 \\
\hybridmamba  & 0.25 & w/  & 0.26 & 0.27 & 0.25 & 0.01 & 0.51 & 0.50 \\
\hybridmamba  & 0.50 & w/o & 0.23 & 0.24 & 0.26 & 0.02 & 0.50 & 0.49 \\
\hybridmamba  & 0.50 & w/  & 0.17 & 0.35 & 0.27 & 0.06 & 0.56 & 0.51 \\
\mambab-2.7B  & 0.25 & w/o & 0.30 & 0.59 & 0.37 & 0.32 & 0.7 & 0.59 \\
\mambab-2.7B  & 0.25 & w/  & 0.31 & 0.59 & 0.38 & 0.32 & 0.69 & 0.59 \\
\mambab-2.7B  & 0.50 & w/o & 0.30 & 0.57 & 0.37 & 0.32 & 0.69 & 0.59 \\
\mambab-2.7B  & 0.50 & w/  & 0.30 & 0.57 & 0.38 & 0.32 & 0.69 & 0.59 \\
\phimamba & 0.25 & w/o & 0.40 & 0.73 & 0.45 & 0.36 & 0.75 & 0.67 \\
\phimamba & 0.25 & w/  & 0.40 & 0.74 & 0.46 & 0.46 & 0.75 & 0.71 \\
\phimamba & 0.50 & w/o & 0.38 & 0.71 & 0.43 & 0.34 & 0.75 & 0.65 \\
\phimamba & 0.50 & w/  & 0.41 & 0.72 & 0.45 & 0.45 & 0.75 & 0.69 \\
\smol         & 0.25 & w/o & 0.40 & 0.73 & 0.48 & 0.53 & 0.76 & 0.61 \\
\smol         & 0.25 & w/  & 0.42 & 0.74 & 0.50 & 0.55 & 0.77 & 0.62 \\
\smol         & 0.50 & w/o & 0.31 & 0.63 & 0.40 & 0.27 & 0.70 & 0.52 \\
\smol         & 0.50 & w/  & 0.40 & 0.74 & 0.49 & 0.54 & 0.76 & 0.60 \\
\bottomrule
\end{tabular}
\caption{\label{tab:state_results} State pruning benchmarks full results. See \Cref{fig:state_radar} for radar plot visualization.}
\end{table*}

\begin{table*}[ht]
\centering
\small
\begin{tabular}{l c c c c c c c c}
\toprule
\textbf{Model} & \textbf{Ratio} & \textbf{arc\_challenge} & \textbf{arc\_easy} & \textbf{hellaswag} & \textbf{lambada} & \textbf{piqa} & \textbf{winogrande} & \textbf{Average} \\
\midrule
\smol         & 0.50 & 0.37 & 0.72 & 0.47 & 0.42 & 0.76 & 0.59 & 0.56 \\
\smol         & 0.25 & 0.42 & 0.75 & 0.50 & 0.56 & 0.77 & 0.63 & 0.61 \\
\smol         & 0    & 0.43 & 0.75 & 0.51 & 0.57 & 0.77 & 0.63 & 0.61 \\
\bottomrule
\end{tabular}
\caption[]{\label{tab:wanda_smol} WANDA pruning on SMOL models. With 25\% sparsity both transformer and Mamba variants show negligible loss; at 50\% the transformer drops $4\%$ while the Mamba variant (\smol) drops $\sim$4\%. See \Cref{fig:wanda_comparison_radar} for radar plot visualization.}
\label{tab:wanda_vs_tr}
\end{table*}

\begin{table*}[ht]
\centering
\small
\begin{tabular}{c c c c c}
\toprule
\textbf{Batch} & \textbf{Seq} & \textbf{FLAP\,p0.5} & \textbf{FLAP\,p0.25} & \textbf{\smol} \\
\textbf{Size} & \textbf{Len.} & \textbf{(tokens/s)} & \textbf{(tokens/s)} & \textbf{(tokens/s)} \\
\midrule
1  & 128  & 3328.13 (x0.99) & 3341.27 (x1.00) & 3346.05 (x1.00) \\
1  & 512  & 9110.18 (x1.15) & 8284.75 (x1.04) & 7925.90 (x1.00) \\
1  & 1024 & 10315.85 (x1.05) & 9992.27 (x1.01) & 9866.42 (x1.00) \\
4  & 128  & 9012.99 (x1.17) & 8145.97 (x1.06) & 7668.75 (x1.00) \\
4  & 512  & 12415.47 (x1.13) & 11319.67 (x1.03) & 10997.28 (x1.00) \\
4  & 1024 & 13007.35 (x1.12) & 11682.80 (x1.00) & 11650.94 (x1.00) \\
16 & 128  & 12395.10 (x1.16) & 11284.14 (x1.06) & 10647.40 (x1.00) \\
16 & 512  & 13635.86 (x1.14) & 12373.93 (x1.04) & 11931.12 (x1.00) \\
16 & 1024 & 14216.09 (x1.10) & 12733.21 (x0.98) & 12964.43 (x1.00) \\
\bottomrule
\end{tabular}
\caption[]{Batch-inference throughput (tokens / s) on a single NVIDIA A100-40GB (128 GB RAM, 128 CPU cores). FLAP models are pruned only in the Mamba component; figures in parentheses show speed-up over the baseline. Pruning 50\% yields 5–17\% gains, while 25\% pruning gives 0–6\%.}
\label{tab:flap_throughput}
\end{table*}

\begin{table*}[ht]
\centering
\small
\begin{tabular}{l c c c c c c c}
\toprule
Model         & Ratio & arc\_challenge & arc\_easy & hellaswag & lambada & piqa & winogrande \\
\midrule
LLAMBA-1B     & 0.50 & 0.32 & 0.68 & 0.43 & 0.43 & 0.73 & 0.58 \\
LLAMBA-1B     & 0.25 & 0.33 & 0.69 & 0.45 & 0.48 & 0.74 & 0.60 \\
LLAMBA-1B     & 0    & 0.33 & 0.70 & 0.45 & 0.49 & 0.74 & 0.61 \\
\bottomrule
\end{tabular}
\caption{WANDA pruning benchmarks results for LLAMBA 1B. See \Cref{fig:llamba_wanda_radar} for radar plot visualization.}
\label{tab:wanda_results_llamba}
\end{table*}

\begin{table*}[ht]
\centering
\small
\begin{tabular}{l c c c c c c c}
\toprule
\textbf{Model}    & \textbf{Ratio} & \textbf{arc\_challenge} & \textbf{arc\_easy} & \textbf{hellaswag} & \textbf{lambada} & \textbf{piqa} & \textbf{winogrande} \\
\midrule
LLAMBA-1B         & 0.25 & 0.27 & 0.64 & 0.37 & 0.27 & 0.70 & 0.52 \\
LLAMBA-1B         & 0.50 & 0.18 & 0.37 & 0.28 & 0.01 & 0.60 & 0.51 \\
\bottomrule
\end{tabular}
\caption{FLAP benchmark results for LLAMBA 1B. See \Cref{fig:llamba_flap_radar} for radar plot visualization.}
\label{tab:flap_results_llamba}
\end{table*}

\FloatBarrier
\clearpage
\section{Radar Plot Visualizations}
\label{app:radar_plots}

This section contains radar plot visualizations corresponding to the results tables in the appendix.

\begin{figure}[ht]
\centering
\begin{tikzpicture}
\small
\begin{polaraxis}[
    width=5.2cm, height=5.5cm,
    xtick={0,60,120,180,240,300},
    xticklabels={ARC-C, ARC-E, HSwag, LAMBADA, PIQA, Wino},
    ytick={0,0.2,0.4,0.6,0.8},
    ymin=0, ymax=0.8,
    grid=both,
    title={\textbf{Head Dimension Pruning - \phimamba}},
    legend pos=south east,
    legend style={font=\tiny}
]
\addplot[thick, blue, mark=o, mark size=2pt] coordinates {
    (0,0.41) (60,0.74) (120,0.46) (180,0.50) (240,0.76) (300,0.72) (0,0.41)
};
\addlegendentry{Dense}

\addplot[thick, red, mark=square, mark size=2pt] coordinates {
    (0,0.36) (60,0.70) (120,0.42) (180,0.35) (240,0.74) (300,0.63) (0,0.36)
};
\addlegendentry{25\% w/ FT}

\addplot[thick, green, mark=triangle, mark size=2pt] coordinates {
    (0,0.29) (60,0.65) (120,0.38) (180,0.22) (240,0.71) (300,0.54) (0,0.29)
};
\addlegendentry{50\% w/ FT}
\end{polaraxis}
\end{tikzpicture}
\hfill
\begin{tikzpicture}
\small
\begin{polaraxis}[
    width=5.2cm, height=5.5cm,
    xtick={0,60,120,180,240,300},
    xticklabels={ARC-C, ARC-E, HSwag, LAMBADA, PIQA, Wino},
    ytick={0,0.2,0.4,0.6,0.8},
    ymin=0, ymax=0.8,
    grid=both,
    title={\textbf{Head Dimension Pruning - \hybridmamba}},
    legend pos=south east,
    legend style={font=\tiny}
]
\addplot[thick, blue, mark=o, mark size=2pt] coordinates {
    (0,0.51) (60,0.80) (120,0.55) (180,0.55) (240,0.77) (300,0.67) (0,0.51)
};
\addlegendentry{Dense}

\addplot[thick, red, mark=square, mark size=2pt] coordinates {
    (0,0.20) (60,0.34) (120,0.27) (180,0.06) (240,0.56) (300,0.51) (0,0.20)
};
\addlegendentry{25\% w/ FT}

\addplot[thick, green, mark=triangle, mark size=2pt] coordinates {
    (0,0.17) (60,0.35) (120,0.27) (180,0.06) (240,0.56) (300,0.51) (0,0.17)
};
\addlegendentry{50\% w/ FT}
\end{polaraxis}
\end{tikzpicture}
\caption{Radar plots showing the effect of head dimension pruning ratios on \phimamba and \hybridmamba across all benchmarks.}
\label{fig:headdim_radar}
\end{figure}

\begin{figure}[ht]
\centering
\begin{tikzpicture}
\small
\begin{polaraxis}[
    width=4.2cm, height=4.5cm,
    xtick={0,60,120,180,240,300},
    xticklabels={ARC-C, ARC-E, HSwag, LAMBADA, PIQA, Wino},
    ytick={0,0.2,0.4,0.6,0.8},
    ymin=0, ymax=0.8,
    grid=both,
    title={\textbf{WANDA Pruning - \phimamba}},
    legend pos=south east,
    legend style={font=\scriptsize}
]
\addplot[thick, blue, mark=o, mark size=2pt] coordinates {
    (0,0.41) (60,0.74) (120,0.46) (180,0.50) (240,0.76) (300,0.72) (0,0.41)
};
\addlegendentry{Dense}

\addplot[thick, red, mark=square, mark size=2pt] coordinates {
    (0,0.41) (60,0.74) (120,0.46) (180,0.50) (240,0.76) (300,0.72) (0,0.41)
};
\addlegendentry{25\% WANDA}

\addplot[thick, green, mark=triangle, mark size=2pt] coordinates {
    (0,0.39) (60,0.72) (120,0.45) (180,0.40) (240,0.75) (300,0.69) (0,0.39)
};
\addlegendentry{50\% WANDA}
\end{polaraxis}
\end{tikzpicture}
\hfill
\begin{tikzpicture}
\small
\begin{polaraxis}[
    width=4.2cm, height=4.5cm,
    xtick={0,60,120,180,240,300},
    xticklabels={ARC-C, ARC-E, HSwag, LAMBADA, PIQA, Wino},
    ytick={0,0.2,0.4,0.6,0.8},
    ymin=0, ymax=0.8,
    grid=both,
    title={\textbf{WANDA Pruning - \hybridmamba}},
    legend pos=south east,
    legend style={font=\scriptsize}
]
\addplot[thick, blue, mark=o, mark size=2pt] coordinates {
    (0,0.51) (60,0.80) (120,0.55) (180,0.55) (240,0.77) (300,0.67) (0,0.51)
};
\addlegendentry{Dense}

\addplot[thick, red, mark=square, mark size=2pt] coordinates {
    (0,0.51) (60,0.80) (120,0.55) (180,0.55) (240,0.77) (300,0.68) (0,0.51)
};
\addlegendentry{25\% WANDA}

\addplot[thick, green, mark=triangle, mark size=2pt] coordinates {
    (0,0.51) (60,0.80) (120,0.54) (180,0.54) (240,0.76) (300,0.66) (0,0.51)
};
\addlegendentry{50\% WANDA}
\end{polaraxis}
\end{tikzpicture}

\begin{tikzpicture}
\small
\begin{polaraxis}[
    width=4.2cm, height=4.5cm,
    xtick={0,60,120,180,240,300},
    xticklabels={ARC-C, ARC-E, HSwag, LAMBADA, PIQA, Wino},
    ytick={0,0.2,0.4,0.6,0.8},
    ymin=0, ymax=0.8,
    grid=both,
    title={\textbf{WANDA Pruning - \smol}},
    legend pos=south east,
    legend style={font=\scriptsize}
]
\addplot[thick, blue, mark=o, mark size=2pt] coordinates {
    (0,0.43) (60,0.75) (120,0.51) (180,0.57) (240,0.77) (300,0.63) (0,0.43)
};
\addlegendentry{Dense}

\addplot[thick, red, mark=square, mark size=2pt] coordinates {
    (0,0.42) (60,0.75) (120,0.50) (180,0.56) (240,0.77) (300,0.63) (0,0.42)
};
\addlegendentry{25\% WANDA}

\addplot[thick, green, mark=triangle, mark size=2pt] coordinates {
    (0,0.37) (60,0.72) (120,0.47) (180,0.42) (240,0.76) (300,0.59) (0,0.37)
};
\addlegendentry{50\% WANDA}
\end{polaraxis}
\end{tikzpicture}
\hfill
\begin{tikzpicture}
\small
\begin{polaraxis}[
    width=4.2cm, height=4.5cm,
    xtick={0,60,120,180,240,300},
    xticklabels={ARC-C, ARC-E, HSwag, LAMBADA, PIQA, Wino},
    ytick={0,0.2,0.4,0.6,0.8},
    ymin=0, ymax=0.8,
    grid=both,
    title={\textbf{WANDA Pruning - \mambab-2.7B}},
    legend pos=south east,
    legend style={font=\scriptsize}
]
\addplot[thick, blue, mark=o, mark size=2pt] coordinates {
    (0,0.33) (60,0.70) (120,0.50) (180,0.70) (240,0.76) (300,0.64) (0,0.33)
};
\addlegendentry{Dense}

\addplot[thick, red, mark=square, mark size=2pt] coordinates {
    (0,0.31) (60,0.63) (120,0.42) (180,0.49) (240,0.72) (300,0.58) (0,0.31)
};
\addlegendentry{25\% WANDA}

\addplot[thick, green, mark=triangle, mark size=2pt] coordinates {
    (0,0.20) (60,0.35) (120,0.28) (180,0.02) (240,0.57) (300,0.54) (0,0.20)
};
\addlegendentry{50\% WANDA}
\end{polaraxis}
\end{tikzpicture}
\caption{Radar plots showing the effect of WANDA pruning ratios on all four models across all benchmarks.}
\label{fig:wanda_radar}
\end{figure}

\begin{figure}[ht]
\centering
\begin{tikzpicture}
\small
\begin{polaraxis}[
    width=4.2cm, height=4.5cm,
    xtick={0,60,120,180,240,300},
    xticklabels={ARC-C, ARC-E, HSwag, LAMBADA, PIQA, Wino},
    ytick={0,0.2,0.4,0.6,0.8},
    ymin=0, ymax=0.8,
    grid=both,
    title={\textbf{Head Merging - \phimamba}},
    legend pos=south east,
    legend style={font=\scriptsize}
]
\addplot[thick, blue, mark=o, mark size=2pt] coordinates {
    (0,0.41) (60,0.74) (120,0.46) (180,0.50) (240,0.76) (300,0.72) (0,0.41)
};
\addlegendentry{32 heads (Dense)}

\addplot[thick, red, mark=square, mark size=2pt] coordinates {
    (0,0.45) (60,0.73) (120,0.49) (180,0.43) (240,0.76) (300,0.62) (0,0.45)
};
\addlegendentry{16 heads w/ FT}

\addplot[thick, green, mark=triangle, mark size=2pt] coordinates {
    (0,0.41) (60,0.72) (120,0.47) (180,0.42) (240,0.75) (300,0.58) (0,0.41)
};
\addlegendentry{8 heads w/ FT}
\end{polaraxis}
\end{tikzpicture}
\hfill
\begin{tikzpicture}
\small
\begin{polaraxis}[
    width=4.2cm, height=4.5cm,
    xtick={0,60,120,180,240,300},
    xticklabels={ARC-C, ARC-E, HSwag, LAMBADA, PIQA, Wino},
    ytick={0,0.2,0.4,0.6,0.8},
    ymin=0, ymax=0.8,
    grid=both,
    title={\textbf{Head Merging - \smol}},
    legend pos=south east,
    legend style={font=\scriptsize}
]
\addplot[thick, blue, mark=o, mark size=2pt] coordinates {
    (0,0.43) (60,0.75) (120,0.51) (180,0.57) (240,0.77) (300,0.63) (0,0.43)
};
\addlegendentry{32 heads (Dense)}

\addplot[thick, red, mark=square, mark size=2pt] coordinates {
    (0,0.27) (60,0.59) (120,0.41) (180,0.14) (240,0.70) (300,0.53) (0,0.27)
};
\addlegendentry{16 heads w/ FT}

\addplot[thick, green, mark=triangle, mark size=2pt] coordinates {
    (0,0.32) (60,0.66) (120,0.40) (180,0.25) (240,0.72) (300,0.54) (0,0.32)
};
\addlegendentry{8 heads w/ FT}
\end{polaraxis}
\end{tikzpicture}

\begin{tikzpicture}
\small
\begin{polaraxis}[
    width=4.5cm, height=4.5cm,
    xtick={0,60,120,180,240,300},
    xticklabels={ARC-C, ARC-E, HSwag, LAMBADA, PIQA, Wino},
    ytick={0,0.2,0.4,0.6,0.8},
    ymin=0, ymax=0.8,
    grid=both,
    title={\textbf{Head Merging - \mambab-2.7B}},
    legend pos=outer north east,
    legend style={font=\scriptsize}
]
\addplot[thick, blue, mark=o, mark size=2pt] coordinates {
    (0,0.33) (60,0.70) (120,0.50) (180,0.70) (240,0.76) (300,0.64) (0,0.33)
};
\addlegendentry{80 heads (Dense)}

\addplot[thick, red, mark=square, mark size=2pt] coordinates {
    (0,0.21) (60,0.27) (120,0.25) (180,0.00) (240,0.52) (300,0.50) (0,0.21)
};
\addlegendentry{40 heads w/ FT}

\addplot[thick, green, mark=triangle, mark size=2pt] coordinates {
    (0,0.21) (60,0.26) (120,0.26) (180,0.00) (240,0.52) (300,0.48) (0,0.21)
};
\addlegendentry{20 heads w/ FT}
\end{polaraxis}
\end{tikzpicture}
\hfill
\begin{tikzpicture}
\small
\begin{polaraxis}[
    width=4.5cm, height=4.5cm,
    xtick={0,60,120,180,240,300},
    xticklabels={ARC-C, ARC-E, HSwag, LAMBADA, PIQA, Wino},
    ytick={0,0.2,0.4,0.6,0.8},
    ymin=0, ymax=0.8,
    grid=both,
    title={\textbf{Head Merging - \hybridmamba}},
    legend pos=outer north east,
    legend style={font=\scriptsize}
]
\addplot[thick, blue, mark=o, mark size=2pt] coordinates {
    (0,0.51) (60,0.80) (120,0.55) (180,0.55) (240,0.77) (300,0.67) (0,0.51)
};
\addlegendentry{8 heads (Dense)}

\addplot[thick, red, mark=square, mark size=2pt] coordinates {
    (0,0.22) (60,0.27) (120,0.26) (180,0.00) (240,0.53) (300,0.50) (0,0.22)
};
\addlegendentry{4 heads w/ FT}

\addplot[thick, green, mark=triangle, mark size=2pt] coordinates {
    (0,0.23) (60,0.28) (120,0.27) (180,0.00) (240,0.54) (300,0.49) (0,0.23)
};
\addlegendentry{2 heads w/ FT}
\end{polaraxis}
\end{tikzpicture}
\caption{Radar plots showing the effect of head merging on all four models across all benchmarks.}
\label{fig:merge_radar}
\end{figure}

\begin{figure}[ht]
\centering
\begin{tikzpicture}
\small
\begin{polaraxis}[
    width=4.5cm, height=4.5cm,
    xtick={0,60,120,180,240,300},
    xticklabels={ARC-C, ARC-E, HSwag, LAMBADA, PIQA, Wino},
    ytick={0,0.2,0.4,0.6,0.8},
    ymin=0, ymax=0.8,
    grid=both,
    title={\textbf{State Pruning - \phimamba}},
    legend pos=outer north east,
    legend style={font=\scriptsize}
]
\addplot[thick, blue, mark=o, mark size=2pt] coordinates {
    (0,0.41) (60,0.74) (120,0.46) (180,0.50) (240,0.76) (300,0.72) (0,0.41)
};
\addlegendentry{Dense}

\addplot[thick, red, mark=square, mark size=2pt] coordinates {
    (0,0.40) (60,0.74) (120,0.46) (180,0.46) (240,0.75) (300,0.71) (0,0.40)
};
\addlegendentry{25\% w/ FT}

\addplot[thick, green, mark=triangle, mark size=2pt] coordinates {
    (0,0.41) (60,0.72) (120,0.45) (180,0.45) (240,0.75) (300,0.69) (0,0.41)
};
\addlegendentry{50\% w/ FT}
\end{polaraxis}
\end{tikzpicture}
\hfill
\begin{tikzpicture}
\small
\begin{polaraxis}[
    width=4.5cm, height=4.5cm,
    xtick={0,60,120,180,240,300},
    xticklabels={ARC-C, ARC-E, HSwag, LAMBADA, PIQA, Wino},
    ytick={0,0.2,0.4,0.6,0.8},
    ymin=0, ymax=0.8,
    grid=both,
    title={\textbf{State Pruning - \hybridmamba}},
    legend pos=outer north east,
    legend style={font=\scriptsize}
]
\addplot[thick, blue, mark=o, mark size=2pt] coordinates {
    (0,0.51) (60,0.80) (120,0.55) (180,0.55) (240,0.77) (300,0.67) (0,0.51)
};
\addlegendentry{Dense}

\addplot[thick, red, mark=square, mark size=2pt] coordinates {
    (0,0.26) (60,0.27) (120,0.25) (180,0.01) (240,0.51) (300,0.50) (0,0.26)
};
\addlegendentry{25\% w/ FT}

\addplot[thick, green, mark=triangle, mark size=2pt] coordinates {
    (0,0.17) (60,0.35) (120,0.27) (180,0.06) (240,0.56) (300,0.51) (0,0.17)
};
\addlegendentry{50\% w/ FT}
\end{polaraxis}
\end{tikzpicture}

\begin{tikzpicture}
\small
\begin{polaraxis}[
    width=4.5cm, height=4.5cm,
    xtick={0,60,120,180,240,300},
    xticklabels={ARC-C, ARC-E, HSwag, LAMBADA, PIQA, Wino},
    ytick={0,0.2,0.4,0.6,0.8},
    ymin=0, ymax=0.8,
    grid=both,
    title={\textbf{State Pruning - \smol}},
    legend pos=outer north east,
    legend style={font=\scriptsize}
]
\addplot[thick, blue, mark=o, mark size=2pt] coordinates {
    (0,0.43) (60,0.75) (120,0.51) (180,0.57) (240,0.77) (300,0.63) (0,0.43)
};
\addlegendentry{Dense}

\addplot[thick, red, mark=square, mark size=2pt] coordinates {
    (0,0.42) (60,0.74) (120,0.50) (180,0.55) (240,0.77) (300,0.62) (0,0.42)
};
\addlegendentry{25\% w/ FT}

\addplot[thick, green, mark=triangle, mark size=2pt] coordinates {
    (0,0.40) (60,0.74) (120,0.49) (180,0.54) (240,0.76) (300,0.60) (0,0.40)
};
\addlegendentry{50\% w/ FT}
\end{polaraxis}
\end{tikzpicture}
\hfill
\begin{tikzpicture}
\small
\begin{polaraxis}[
    width=4.5cm, height=4.5cm,
    xtick={0,60,120,180,240,300},
    xticklabels={ARC-C, ARC-E, HSwag, LAMBADA, PIQA, Wino},
    ytick={0,0.2,0.4,0.6,0.8},
    ymin=0, ymax=0.8,
    grid=both,
    title={\textbf{State Pruning - \mambab-2.7B}},
    legend pos=outer north east,
    legend style={font=\scriptsize}
]
\addplot[thick, blue, mark=o, mark size=2pt] coordinates {
    (0,0.33) (60,0.70) (120,0.50) (180,0.70) (240,0.76) (300,0.64) (0,0.33)
};
\addlegendentry{Dense}

\addplot[thick, red, mark=square, mark size=2pt] coordinates {
    (0,0.31) (60,0.59) (120,0.38) (180,0.32) (240,0.69) (300,0.59) (0,0.31)
};
\addlegendentry{25\% w/ FT}

\addplot[thick, green, mark=triangle, mark size=2pt] coordinates {
    (0,0.30) (60,0.57) (120,0.38) (180,0.32) (240,0.69) (300,0.59) (0,0.30)
};
\addlegendentry{50\% w/ FT}
\end{polaraxis}
\end{tikzpicture}
\caption{Radar plots showing the effect of state pruning ratios on all four models across all benchmarks.}
\label{fig:state_radar}
\end{figure}

\begin{figure}[ht]
\centering
\begin{tikzpicture}
\small
\begin{polaraxis}[
    width=7.6cm, height=8cm,
    xtick={0,60,120,180,240,300},
    xticklabels={ARC-C, ARC-E, HSwag, LAMBADA, PIQA, Wino},
    ytick={0,0.2,0.4,0.6,0.8},
    ymin=0, ymax=0.8,
    grid=both,
    title={\textbf{WANDA Pruning: Transformer vs Mamba (\smol)}},
    legend pos=south east,
    legend style={font=\scriptsize}
]
\addplot[thick, blue, mark=o, mark size=3pt] coordinates {
    (0,0.45) (60,0.78) (120,0.54) (180,0.68) (240,0.77) (300,0.66) (0,0.45)
};
\addlegendentry{Smol17 (TF) Dense}

\addplot[thick, red, mark=square, mark size=3pt] coordinates {
    (0,0.44) (60,0.77) (120,0.53) (180,0.67) (240,0.77) (300,0.66) (0,0.44)
};
\addlegendentry{Smol17 (TF) 25\%}

\addplot[thick, green, mark=triangle, mark size=3pt] coordinates {
    (0,0.41) (60,0.75) (120,0.50) (180,0.56) (240,0.76) (300,0.62) (0,0.41)
};
\addlegendentry{Smol17 (TF) 50\%}

\addplot[thick, orange, mark=diamond, mark size=3pt] coordinates {
    (0,0.43) (60,0.75) (120,0.51) (180,0.57) (240,0.77) (300,0.63) (0,0.43)
};
\addlegendentry{\smol Dense}

\addplot[thick, purple, mark=pentagon, mark size=3pt] coordinates {
    (0,0.42) (60,0.75) (120,0.50) (180,0.56) (240,0.77) (300,0.63) (0,0.42)
};
\addlegendentry{\smol 25\%}

\addplot[thick, brown, mark=star, mark size=3pt] coordinates {
    (0,0.37) (60,0.72) (120,0.47) (180,0.42) (240,0.76) (300,0.59) (0,0.37)
};
\addlegendentry{\smol 50\%}
\end{polaraxis}
\end{tikzpicture}
\caption{Radar plot comparing WANDA pruning effects on transformer (Smol17) vs Mamba (\smol) variants across all benchmarks and pruning ratios.}
\label{fig:wanda_comparison_radar}
\end{figure}

\begin{figure}[ht]
\centering
\begin{tikzpicture}
\small
\begin{polaraxis}[
    width=6.7cm, height=7cm,
    xtick={0,60,120,180,240,300},
    xticklabels={ARC-C, ARC-E, HSwag, LAMBADA, PIQA, Wino},
    ytick={0,0.2,0.4,0.6,0.8},
    ymin=0, ymax=0.8,
    grid=both,
    title={\textbf{WANDA Pruning - LLAMBA-1B}},
    legend pos=south east,
    legend style={font=\small}
]
\addplot[thick, blue, mark=o, mark size=3pt] coordinates {
    (0,0.33) (60,0.70) (120,0.45) (180,0.49) (240,0.74) (300,0.61) (0,0.33)
};
\addlegendentry{Dense}

\addplot[thick, red, mark=square, mark size=3pt] coordinates {
    (0,0.33) (60,0.69) (120,0.45) (180,0.48) (240,0.74) (300,0.60) (0,0.33)
};
\addlegendentry{25\% WANDA}

\addplot[thick, green, mark=triangle, mark size=3pt] coordinates {
    (0,0.32) (60,0.68) (120,0.43) (180,0.43) (240,0.73) (300,0.58) (0,0.32)
};
\addlegendentry{50\% WANDA}
\end{polaraxis}
\end{tikzpicture}
\caption{Radar plot showing the effect of WANDA pruning ratios on LLAMBA-1B across all benchmarks.}
\label{fig:llamba_wanda_radar}
\end{figure}

\begin{figure}[ht]
\centering
\begin{tikzpicture}
\small
\begin{polaraxis}[
    width=6.7cm, height=7cm,
    xtick={0,60,120,180,240,300},
    xticklabels={ARC-C, ARC-E, HSwag, LAMBADA, PIQA, Wino},
    ytick={0,0.2,0.4,0.6,0.8},
    ymin=0, ymax=0.8,
    grid=both,
    title={\textbf{FLAP Pruning - LLAMBA-1B}},
    legend pos=south east,
    legend style={font=\small}
]
\addplot[thick, blue, mark=o, mark size=3pt] coordinates {
    (0,0.33) (60,0.70) (120,0.45) (180,0.49) (240,0.74) (300,0.61) (0,0.33)
};
\addlegendentry{Dense}

\addplot[thick, red, mark=square, mark size=3pt] coordinates {
    (0,0.27) (60,0.64) (120,0.37) (180,0.27) (240,0.70) (300,0.52) (0,0.27)
};
\addlegendentry{p0.25}

\addplot[thick, green, mark=triangle, mark size=3pt] coordinates {
    (0,0.18) (60,0.37) (120,0.28) (180,0.01) (240,0.60) (300,0.51) (0,0.18)
};
\addlegendentry{p0.50}
\end{polaraxis}
\end{tikzpicture}
\caption{Radar plot showing the effect of FLAP pruning ratios on LLAMBA-1B across all benchmarks.}
\label{fig:llamba_flap_radar}
\end{figure}

\begin{figure}[ht]
\centering
\begin{tikzpicture}
\small
\begin{polaraxis}[
    width=4.2cm, height=4.5cm,
    xtick={0,60,120,180,240,300},
    xticklabels={ARC-C, ARC-E, HSwag, LAMBADA, PIQA, Wino},
    ytick={0,0.2,0.4,0.6,0.8},
    ymin=0, ymax=0.8,
    grid=both,
    title={\textbf{FLAP Pruning - \phimamba}},
    legend pos=south east,
    legend style={font=\scriptsize}
]
\addplot[thick, blue, mark=o, mark size=2pt] coordinates {
    (0,0.41) (60,0.74) (120,0.46) (180,0.50) (240,0.76) (300,0.72) (0,0.41)
};
\addlegendentry{Dense}

\addplot[thick, red, mark=square, mark size=2pt] coordinates {
    (0,0.39) (60,0.72) (120,0.44) (180,0.44) (240,0.75) (300,0.66) (0,0.39)
};
\addlegendentry{p0.25}

\addplot[thick, green, mark=triangle, mark size=2pt] coordinates {
    (0,0.34) (60,0.69) (120,0.40) (180,0.17) (240,0.74) (300,0.58) (0,0.34)
};
\addlegendentry{p0.50}
\end{polaraxis}
\end{tikzpicture}
\hfill
\begin{tikzpicture}
\small
\begin{polaraxis}[
    width=4.2cm, height=4.5cm,
    xtick={0,60,120,180,240,300},
    xticklabels={ARC-C, ARC-E, HSwag, LAMBADA, PIQA, Wino},
    ytick={0,0.2,0.4,0.6,0.8},
    ymin=0, ymax=0.8,
    grid=both,
    title={\textbf{FLAP Pruning - \hybridmamba}},
    legend pos=south east,
    legend style={font=\scriptsize}
]
\addplot[thick, blue, mark=o, mark size=2pt] coordinates {
    (0,0.51) (60,0.80) (120,0.55) (180,0.55) (240,0.77) (300,0.67) (0,0.51)
};
\addlegendentry{Dense}

\addplot[thick, red, mark=square, mark size=2pt] coordinates {
    (0,0.22) (60,0.27) (120,0.26) (180,0.00) (240,0.52) (300,0.49) (0,0.22)
};
\addlegendentry{p0.25}

\addplot[thick, green, mark=triangle, mark size=2pt] coordinates {
    (0,0.24) (60,0.27) (120,0.26) (180,0.00) (240,0.51) (300,0.45) (0,0.24)
};
\addlegendentry{p0.50}
\end{polaraxis}
\end{tikzpicture}

\begin{tikzpicture}
\small
\begin{polaraxis}[
    width=4.2cm, height=4.5cm,
    xtick={0,60,120,180,240,300},
    xticklabels={ARC-C, ARC-E, HSwag, LAMBADA, PIQA, Wino},
    ytick={0,0.2,0.4,0.6,0.8},
    ymin=0, ymax=0.8,
    grid=both,
    title={\textbf{FLAP Pruning - \smol}},
    legend pos=south east,
    legend style={font=\scriptsize}
]
\addplot[thick, blue, mark=o, mark size=2pt] coordinates {
    (0,0.43) (60,0.75) (120,0.51) (180,0.57) (240,0.77) (300,0.63) (0,0.43)
};
\addlegendentry{Dense}

\addplot[thick, red, mark=square, mark size=2pt] coordinates {
    (0,0.34) (60,0.69) (120,0.44) (180,0.27) (240,0.74) (300,0.55) (0,0.34)
};
\addlegendentry{p0.25}

\addplot[thick, green, mark=triangle, mark size=2pt] coordinates {
    (0,0.26) (60,0.57) (120,0.37) (180,0.06) (240,0.69) (300,0.52) (0,0.26)
};
\addlegendentry{p0.50}
\end{polaraxis}
\end{tikzpicture}
\hfill
\begin{tikzpicture}
\small
\begin{polaraxis}[
    width=4.2cm, height=4.5cm,
    xtick={0,60,120,180,240,300},
    xticklabels={arc\_challenge, arc\_easy, hellaswag, lambada, piqa, winogrande},
    ytick={0,0.2,0.4,0.6,0.8},
    ymin=0, ymax=0.8,
    grid=both,
    title={\textbf{FLAP Pruning - \mambab-2.7B}},
    legend pos=south east,
    legend style={font=\scriptsize}
]
\addplot[thick, blue, mark=o, mark size=2pt] coordinates {
    (0,0.33) (60,0.70) (120,0.50) (180,0.70) (240,0.76) (300,0.64) (0,0.33)
};
\addlegendentry{Dense}

\addplot[thick, red, mark=square, mark size=2pt] coordinates {
    (0,0.23) (60,0.28) (120,0.25) (180,0.00) (240,0.51) (300,0.48) (0,0.23)
};
\addlegendentry{p0.25}

\addplot[thick, green, mark=triangle, mark size=2pt] coordinates {
    (0,0.25) (60,0.28) (120,0.25) (180,0.02) (240,0.50) (300,0.44) (0,0.25)
};
\addlegendentry{p0.50}
\end{polaraxis}
\end{tikzpicture}
\caption{Radar plots showing the effect of FLAP pruning ratios on all four models across all benchmarks.}
\label{fig:flap_radar}
\end{figure}

\end{document}